\documentclass[runningheads]{llncs}

\usepackage{eccv}

\usepackage{eccvabbrv}

\usepackage{booktabs}
\usepackage{graphicx, subcaption, ragged2e}
\usepackage{algorithm}
\usepackage{array}
\usepackage{float}	
\usepackage{color, colortbl}
\usepackage{tabularx}
\usepackage{algpseudocode}
\usepackage{amsmath}
\usepackage{amssymb}

\usepackage[accsupp]{axessibility}  % Improves PDF readability for those with disabilities.

\usepackage{hyperref}

\usepackage{orcidlink}

\graphicspath{ {./images/} }

\begin{document}

\title{Adapting Segment Anything Model to Melanoma Segmentation in Microscopy Slide Images}

\titlerunning{Adapting SAM to Melanoma Segmentation in Microscopy Slide Images}

\author{Qingyuan Liu\inst{1}\orcidlink{0009-0003-3802-6628} \and
Avideh Zakhor\inst{1}\orcidlink{0000-0003-4770-6353}}

\authorrunning{Q.~Liu et al.}

\institute{University of California, Berkeley\\
\email{\{pixelled, avz\}@berkeley.edu}}

\maketitle

\begin{abstract}
  Melanoma segmentation in Whole Slide Images (WSIs) is useful for prognosis and the measurement of crucial prognostic factors such as Breslow depth and primary invasive tumor size. In this paper, we present a novel approach that uses the Segment Anything Model (SAM) for automatic melanoma segmentation in microscopy slide images. Our method employs an initial semantic segmentation model to generate preliminary segmentation masks that are then used to prompt SAM. We design a dynamic prompting strategy that uses a combination of centroid and grid prompts to achieve optimal coverage of the super high-resolution slide images while maintaining the quality of generated prompts. To optimize for invasive melanoma segmentation, we further refine the prompt generation process by implementing in-situ melanoma detection and low-confidence region filtering. We select Segformer as the initial segmentation model and EfficientSAM as the segment anything model for parameter-efficient fine-tuning. Our experimental results demonstrate that this approach not only surpasses other state-of-the-art melanoma segmentation methods but also significantly outperforms the baseline Segformer by 9.1\% in terms of IoU.
\end{abstract}

\section{Introduction}
\label{sec:intro}

Melanoma, one of the most serious forms of skin cancer, originates in melanocytes, the pigment-producing cells responsible for melanin production \cite{skincancer_melanoma}. Based on its progression and location within the skin, melanoma can be categorized into two main types: in-situ melanoma and invasive melanoma. While in-situ melanoma represents cancerous melanocytes that are confined to the epidermis, invasive melanoma penetrates beyond the epidermis into the dermis, posing a significant risk of spreading to other vital organs. As invasive melanoma grows, it may invade blood vessels and lymphatic vessels, allowing cancer cells to detach from the primary tumor and cause metastatic cancer \cite{abeloff2008abeloff}. In this paper, we primarily focus on segmenting invasive melanoma to assist dermatologists in deciding on treatment options.

Early detection and accurate diagnosis of melanoma are crucial for improving the survival rate \cite{acs_cancer_facts_2024, conic2018determination}. The standard diagnosis practice begins with an initial examination of dermatoscopic features to determine the types of melanocyptic lesions. For suspicious and malignant lesions, a histopathologic analysis of skin biopsies stained with hematoxylin and eosin (H\&E) is required \cite{goodson2009strategies}. While traditional approach requires examination of tissue specimens under a microscope, the advent of Whole Slide Images (WSIs) has revolutionized this process by digitizing tissue samples into high resolution images. These digital slide images enable pathologists to examine various characteristics such as the celluar architecture, breslow depth and complex histologic features that are crucial for determining the stage and aggressiveness of melanoma.

Recent studies \cite{nofallah2022segmenting, phillips2019segmentation, alheejawi2020deep, van2020segmentation, oskal2019u, shah2023deep} have focused on utilizing deep learning technology for melanoma segmentation in whole slide images. Melanoma segmentation results with sufficient accuracy in slide images have proven highly beneficial for aiding diagnosis and assisting manual measurement of breslow depth and primary invasive tumor size, which are crucial prognostic factors. This shows the potential for a fully automatic diagnosis procedure. Phillips et al. \cite{phillips2019segmentation} demonstrated the effectiveness of multi-scale FCN in segmentation the dermis, epidermis and tumor in whole slide images. Want et al. \cite{wang2024transformers} utilized more advanced transformer models including Hiearachical Pyramid Transformers (HIPT) and Segformers to achieve accurate segmentation of invasive melanoma and the epidermis in microscopy slide images.

In this paper, we propose a novel method to use the Segment Anything Model (SAM) \cite{kirillov2023segment} for automatic melanoma segmentation in microscopy slide images. SAM has demonstrated great success in various computer vision tasks and has achieved state-of-the-art performance in a diverse range of image segmentation tasks \cite{kirillov2023segment, van2011segmentation, arbelaez2010contour}. One of the key features of SAM is a prompt encoder that allows the model to adapt to diverse downstream segmentation tasks with prompt engineering. Despite SAM's strong zero-shot generalization abilities, several studies \cite{huang2024segment, zhou2023can, roy2023sam, deng2023segment} have shown that its accuracy is limited in segmentation tasks that require specific domain knowledge. Recent studies \cite{ma2024segment, wu2023medical, zhang2023input} have proposed methods to adapt SAM for medical image segmentation, such as augmenting data with SAM's predictions or fine-tuning SAM for better performance. Despite showing impressive performance in segmenting medical images such as CT and MRI scans, these methods do not generalize well to microscopy images, especially for melanoma segmentation in microscopy slide images.

Our proposed method addresses these challenges by introducing an innovative framework that automatically generates prompts from an initial segmentation maps to guide SAM. We use Segformer, an effective semantic segmentation model for melanoma segmentation, to generate the initial segmentation map. We design a dynamic prompt strategy that uses a combination of centroid and grid prompts. Additionally, we incorporate in-situ melanoma detection and low-confidence region filtering to ensure precise prompt generation. Our experimental results demonstrate that this approach not only surpasses other state-of-the-art melanoma segmentation methods but also significantly improves upon the baseline performance of Segformer by over 9.1\%. 

\section{Related Work}

\textbf{SAM in Medical Imaging}. Given SAM's impressive results on various natural image segmentation tasks \cite{kirillov2023segment, arbelaez2010contour, van2011segmentation}, recent works have explored its application to medical image segmentation. However, several studies \cite{huang2024segment, zhou2023can, roy2023sam, deng2023segment} have shown that SAM underperforms in medical image segmentation. This is attributed to the model's lack of domain-specific medical knowledge, the uncertain and complex object boundaries, intricate structures, and the wide-range of scales unique to medical objects \cite{huang2024segment}. Recent efforts have focused on adapting SAM for medical images, primarily through fine-tuning or adapting SAM to labeled medical dataset. Ma et al. \cite{ma2024segment} proposed to fine-tune SAM fully on labeled medical data, which is not cost-effective due to the vast number of parameters. Wu et al. \cite{wu2023medical} proposed to integrate adapter modules into SAM, allowing efficient fine-tuning by freezing all modules except for the adapters during training. These studies all focus on utilizing SAM as an interactive model and evaluate its performance by providing accurate prompts based on the ground-truth annotations. For microscopy slide image segmentation, SAM's interactive feature can assist annotators but cannot fully automate the labor-intensive process of segmenting high-resolution whole slide images. Although SAM offers automatic mask generation, this feature is designed for segmenting everything on an entire image rather than targeting specific small objects, such as scattered melanoma cells in microscopy slide images. Zhang et al. \cite{zhang2023input} proposed to enhance medical images by adding semantic structures using SAM's automatic mask generation. This approach combines generated masks, features and stability scores to help train other image segmentation models with enhanced data. The success of this method depends on SAM's ability to generate useful structural information during the automatic mask generation process.

\noindent \textbf{Melanoma Segmentation in Slide Images}.  Melanoma semantic segmentation can be classified into two categories: 1) skin lesion segmentation based on images captured at a macroscopic scale; 2) microscopy slide image segmentation that involves WSIs captured at a microscopic scale. While skin lesion segmentation only segments a large rounded melanoma component at a macroscopic level, microscopy slide image segmentation requires segmenting irregular melanoma scattered across high-resolution, detailed views of tissue samples at the cellular level, making the task significantly more challenging. Our work focuses exclusively on the latter. 

Most studies \cite{nofallah2022segmenting, phillips2019segmentation, alheejawi2020deep, van2020segmentation, oskal2019u, shah2023deep} focus on utilizing CNN-based methods for melanoma segmentation in microscopy slide images. Phillips et al. \cite{phillips2019segmentation} devised a multi-stride fully convolutional network (FCN) that can effectively segment tumour, epidermis and dermis. Shah et al. \cite{shah2023deep} developed a two-stage method to segment invasive melanoma by leveraging the difference between in-situ melanoma and invasive melanoma. They used HRNet-OCR \cite{yuan2020object} and HookNet \cite{van2021hooknet} as backbones and trained two models, one for epidermis segmentation and one for tumor segmentation, to predict two separate segmentation masks for epidermis and melanoma. By removing all predicted melanoma from the epidermis mask, they obtained segmentation masks for invasive melanoma.

Recent works have applied transformer-based architectures \cite{vaswani2017attention} to melanoma segmentation. Wang et al. \cite{wang2024transformers} proposed to fine-tune the Hierarchical Pyramid Transformer (HIPT) \cite{chen2022adaptformer} and the Segformer \cite{xie2021segformer} for invasive melanoma segmentation in microscopy slide images. Segformer has been shown to outperform all other CNN-based methods and HIPT in segmenting invasive melanoma, achieving state-of-the-art IoU \cite{wang2024transformers}. Our method is designed with Segformer \cite{xie2021segformer} as the initial segmentation model to achieve the best results.

\section{Method}

\begin{figure}[tb]
\centering
\includegraphics[width=\textwidth]{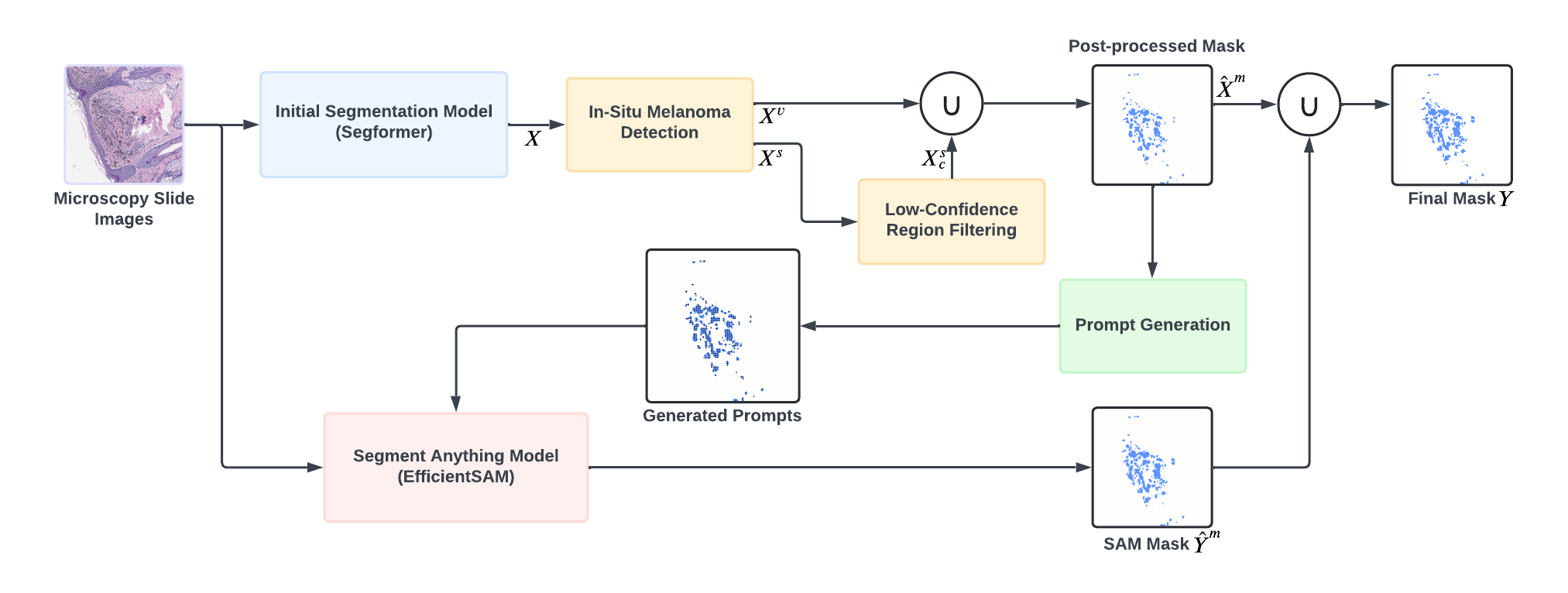}
\caption{An overview of the proposed method. The initial mask $X$ generated by Segformer is post-processed to generate the mask $\hat{X}^m$. We run SAM on the prompts generated from $\hat{X}^m$ to generate the mask $\hat{Y}^m$. The two masks are combined to create the final mask Y.}
\label{fig:framework}
\end{figure}

\subsection{Overview of the Proposed Method \label{sec:overview}}
Our proposed method comprises an initial semantic segmentation model and a segment anything model as illustrated in \cref{fig:framework}. To segment invasive melanoma automatically, we prompt SAM with prompts generated from the mask produced by the initial segmentation model. To optimize for our task of segmenting invasive melanoma, we select Segformer as the initial segmentation model due to its superior performance compared to other models \cite{wang2024transformers}. We choose EfficientSAM \cite{xiong2023efficientsam} as the segment anything model for parameter-efficient fine-tuning. The entire framework is described as follows.

\textbf{Step 1. Initial Mask Generation}. We run Segformer\cite{xie2021segformer} to generate the initial segmentation mask $X$. Then we perform in-situ melanoma detection to separate $X$ into the estimated in-situ melanoma regions $X^s$ and the remaining invasive melanoma regions $X^v$. We filter out low-confidence regions from $X_s$ to obtain the filtered mask $X^s_c$ and combine it with $X^v$ to obtain the post-processed mask $\hat{X}^m$. The details are described in \cref{sec:detection,sec:filter}.

\textbf{Step 2. Prompt Generation}. We generate single point prompts from the post-processed mask $\hat{X}^m$. We determine the best prompt type for each connected component based on its shape distribution and geometric characteristics. The details of the prompt generation strategy are described in \cref{sec:prompt}.

\textbf{Step 3. Final Mask Generation}. We run SAM on the generated prompts to produce its own invasive melanoma mask $\hat{Y}^m$. We use SAM's mask to refine the post-processed mask $\hat{X}^m$ by combining the two masks together. This aims to enhance the overall accuracy and robustness of melanoma segmentation. The details are described in \cref{sec:final}.

\begin{figure}[t]
\captionsetup[subfigure]{justification=Centering}
\begin{subfigure}[t]{0.55\textwidth}
    \centering
    \includegraphics[width=\linewidth]{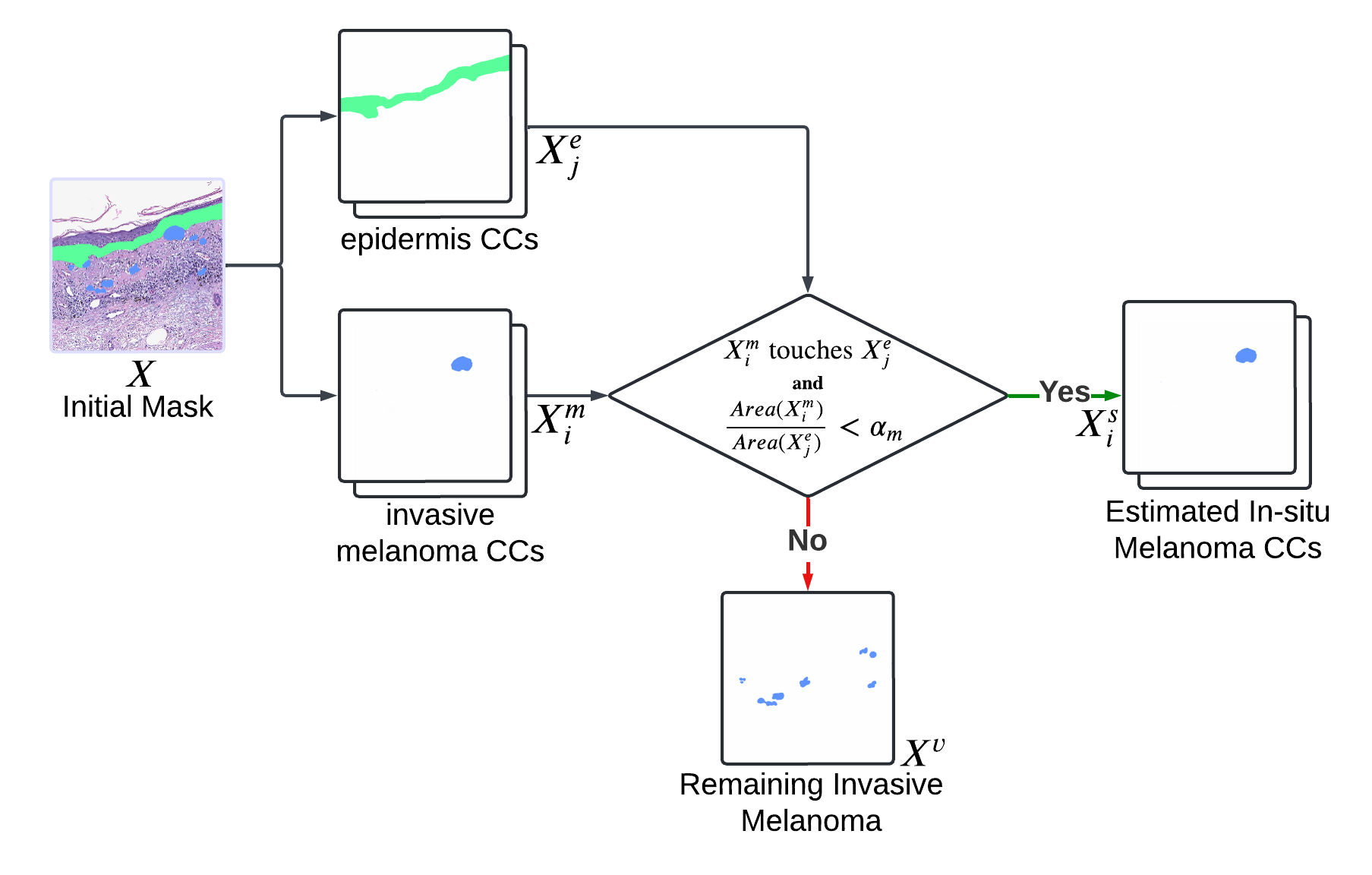}
    \caption{In-situ Melanoma Detection}
    \label{fig:detection}
\end{subfigure}
\centering
\begin{subfigure}[t]{0.35\textwidth}
    \centering
    \includegraphics[width=\linewidth]{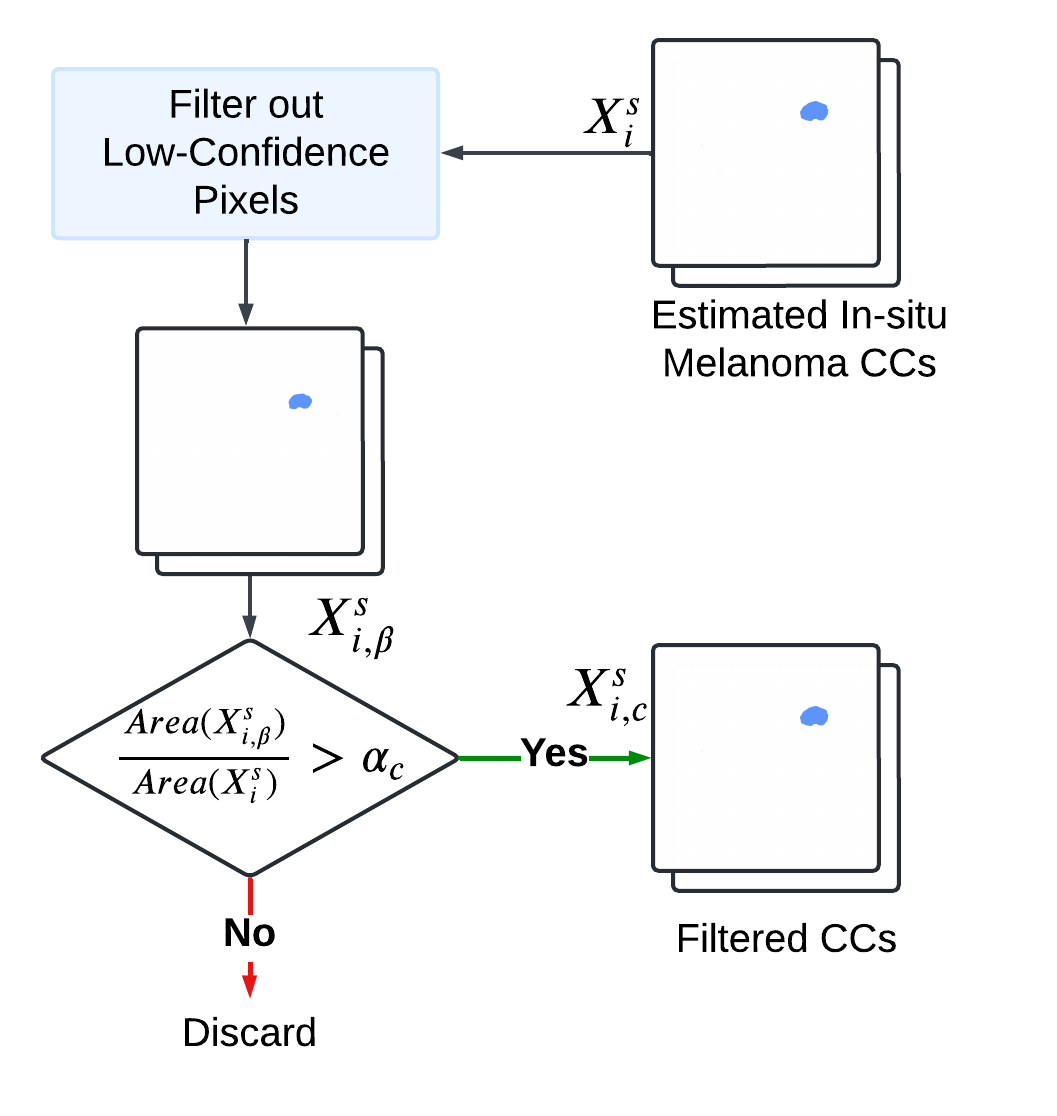}
    \caption{Low-Confidence Region Filtering}
    \label{fig:filter}
\end{subfigure}
\caption{In-situ melanoma detection finds the estimated in-situ melanoma regions $X^s$ from the initial mask $X$. Low-confidence region filtering discards low-confidence connected components from $X^s$.}
\end{figure}

\subsection{In-situ Melanoma Detection \label{sec:detection}}
We exclude low-confidence invasive melanoma predictions that touch the epidermis, since these are considered to be in-situ melanoma from a medical perspective. This process is shown in \cref{fig:detection}. The initial mask $X$ produced by Segformer contains both the epidermis mask $X^e$, and the invasive melanoma mask $X^m$. The invasive melanoma mask can be represented as a union of invasive melanoma connected components (CCs) $X^m = \bigcup X^m_i$, where $i$ denotes the $i$th invasive melanoma connected component. Similarly, the mask for epidermis can be represented as $X^e = \bigcup X^e_j$, where $j$ denotes the $jth$ epidermis connected component. Since SAM heavily relies on prompts to specify the exact objects to segment, inaccurate and ambiguous prompts can significantly degrade its performance. This can lead SAM to produce more false positives compared to the initial segmentation mask from which prompts are generated. Therefore, we filter out low-confidence invasive melanoma components to improve the accuracy of prompts so that it is clicked on true positives in most cases. 

To begin with, we iterate over all connected components $X^m_i$ in the invasive melanoma mask $X^m$. For each connected component, we determine whether it touches any epidermis component $X^e_j$. If such a touch exists, we compute the ratio between the area of the melanoma component $X^m_i$ and the area of the touched epidermis region $X^e_j$. If this ratio falls below a specified threshold $\alpha_m$, we classify it as an estimated in-situ melanoma region, This is because careful inspection of Segformer's predictions shows that in-situ melanoma that are misclassified as invasive melanoma tend to have a relatively small size compared to the epidermis. This process results in the estimated in-situ melanoma mask $X^s$ and the remaining invasive melanoma regions $X^v$.

\subsection{Low-Confidence Region Filtering \label{sec:filter}}
We further filter out low-confidence estimated in-situ melanoma regions from $X^s$. We keep high-confidence regions even if they touch the epidermis, since this could result from false predictions in the epidermis that make them touch each other. We determine the confidence level based on the probability map generated by Segformer. As shown in \cref{fig:filter}, for each component $X^s_i$, we first find its high-confidence sub-component $X^s_{i, \beta}$ that has a probability exceeding the defined threshold $\beta$:
\begin{equation}
    X^s_{i, \beta} = \{x \mid P(x) > \beta, x \in X^s_i\}.
\end{equation}
We then determine the confidence level by computing the ratio between the area of $X^s_{i, \beta}$ and $X^s_i$. If the ratio fails to meet the confidence threshold $\alpha_c$, we discard this component and do not generate prompts from it. Careful inspections of Segformer's mask show that it generally predicts a larger proportion of low-probability areas for in-situ melanoma compared to invasive melanoma. Therefore, if we keep only regions with a high proportion of high probability areas, we can significantly reduce the number of false positives in $X^m$. Next, we combine the filtered melanoma regions $X^s_c$ and the rest invasive melanoma regions $X^v$ to obtain the post-processed invasive melanoma mask $\hat{X}^m$.

\begin{figure}[tb]
\centering
\includegraphics[width=\textwidth]{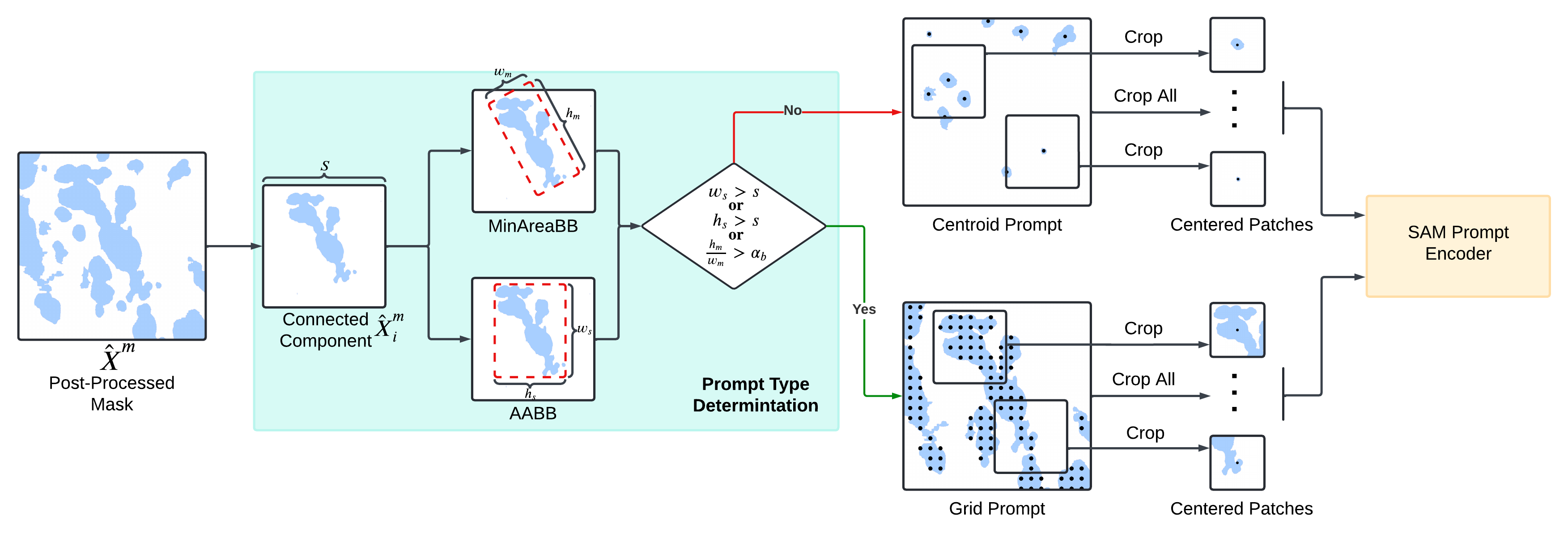}
\caption{Prompt Generation Strategy. We apply the centroid prompt and the grid prompt to based on the geometric characteristics of each connected component. }
\label{fig:prompt}
\end{figure}

\begin{figure}[tb]
\begin{algorithm}[H]
\caption{Determine prompt type for a connected component}
\begin{algorithmic}[1]
\Require
      \Statex A connected component of invasive melanoma: $x$
      \Statex The length of the side of a patch: $s$
      \Statex The threshold for the ratio between sides of a minimum bounding box: $\alpha_b$
\Ensure
    \Statex The generated prompts (a set of single point prompts): $P$
\Function {GeneratePrompts}{x}
    \State $P = \emptyset$
    \State $w_s, h_s = \Call{FindAABB}{x}$
    \State $w_m, h_m = \Call{FindMinAreaBB}{x}$
    \If {$w_s > s$ \textbf{or} $h_s > s$ \textbf{or} $\frac{h_m}{w_m} > \alpha_b$}
        \State $P_{grid} = \Call{GenerateGridPrompts}{x}$
        \State $P = P \cup P_{grid}$
        \State $center = \Call{FindCentroid}{x}$
        \If {$center \in x$}
            \State $P = P \cup \{center\}$
        \EndIf
    \Else
        \State $center = \Call{FindCentroid}{x}$
        \State $P = P \cup \{center\}$
    \EndIf
    \State \Return P
\EndFunction
\end{algorithmic}
\label{alg:prompts}
\end{algorithm}
\end{figure}

\subsection{Prompt Generation \label{sec:prompt}}
Melanoma can be in various forms in microscopy slide images, ranging from rounded clusters to jagged streaks or any combination of irregular shapes. To facilitate accurate segmentation of diverse morphologies, we design two types of point prompts: centroid and grid. Centroid prompts consist of a single point placed at the centroid of a connected component. On the other hand, grid prompts comprise a grid of points distributed within a connected component. As shown in \cref{fig:prompt}, we choose a prompt type for each connected component $\hat{X}^m_i$ and serve each point as a single point prompt for the SAM prompt encoder. We feed a patch centered at the point to the SAM image encoder, since this ideally provides maximum context.

We choose prompts dynamically based on the characteristics of the connected components. Centroid prompts excel in providing optimal context for small, regularly shaped melanoma, but often fail to provide sufficient context for those covering large regions and those with irregular shapes. In cases where a connected component is too large to fit inside a patch, numerous melanoma predictions are lost in a centered patch and not enough context is provided for its surroundings. Furthermore, if the centroid falls outside the component, it turns out to be a point prompt not clicked on invasive melanoma, resulting in false positives.

For each connected component $\hat{X}^m_i$, we determine the prompt type based on its shape distribution and geometric attributes, as detailed in \cref{alg:prompts}. As shown in \cref{fig:prompt}, we find the width $w_s$ and height $h_s$ of an axis aligned bounding box (AABB). If either dimension exceeds the patch length, we use grid prompts. Then we find the width $w_m$ and height $h_m$ of an arbitrarily oriented bounding box with minimum area and check whether the ratio between $h_m$ and $w_m$ exceeds a threshold $\alpha_b$ assuming $h_m$ is the longer side. A high ratio implies that the shape of the component resembles a narrow rectangle or any other irregular shape where one dimension is larger than the other, in which case centroid prompt is ambiguous for segmenting the whole component. In such cases, we also choose to use grid prompts. Additionally, we include its centroid as part of the prompt if it is inside the component as a supplementary to the grid prompts. For all other cases, we use centroid prompt only.

If a connected component that meets the criteria for grid prompts has more than $\frac{1}{4}$ of the patch area, we apply grid prompts to all connected components in the slide image instead of evaluating each one individually. This is because neighboring melanomas often share similar shapes, even if they don't fit a single morphology. When a large melanoma with an irregular shape is present, nearby melanomas usually have similar shapes, making grid prompts the most effective for covering all components.

\subsection{Final Mask Generation \label{sec:final}}
After deriving the prompts, we run SAM to generate its own mask $\hat{Y}^m$ as shown in \cref{fig:framework}. To obtain the final mask, we combine the post-processed Segformer's mask $\hat{X}^m$ and SAM's mask $\hat{Y}^m$ by performing a union between the two:
\begin{equation}
    Y = \hat{Y}^m \cup \hat{X}^m
\end{equation}
We choose to combine these two masks because single point prompts are sparse and may not comprehensively cover all regions. This sparsity can lead to ambiguity and result in missing some high-confidence predictions from Segformer. Therefore, we choose to include the post-processed mask $\hat{X}^m$ to ensure that high-confidence regions are kept in the final mask $Y$.

\subsection{Training Strategy} \label{sec:training}
We devise a training strategy involving two stages that optimizes SAM for the way we use prompts. Since we only use single point prompts, we first train SAM on single point prompts for melanoma. For each patch sampled from a microscopy slide image, we first find all connected components and for each connected component we generate a single point prompt clicked at a random position inside it. This enables the model to learn to segment melanoma anywhere in a patch. After the initial training on random point prompts, we fine-tune our model on patches centered at each individual single point prompt. Each patch might contain multiple components, but we only set the component containing the point prompt as the target to reduce ambiguity. This allows the model to optimize for inference using our SAM-based method.

\section{Experiments}

\subsection{Dataset \label{sec:dataset}}

Our dataset comprises 101 microscopy slide images, with sizes ranging from $23700 \times 21199$ pixels to $1996 \times 1679$ pixels. These images are derived from skin biopsies stained with H\&E \cite{fischer2008hematoxylin} and captured under a microscope at 40x magnification. Detailed Annotations at this magnification level are provided by an expert dermatopathologist. Each image in the dataset is carefully selected and cropped from the original WSIs by the dermatopathologist. This is because whole slide images often contain multiple focal planes of the same tissue, leading to redundancy in the data. By selecting the most informative regions at a single focal plane, we reduce the redundancy and ensure that the dataset focuses on the most relevant tissue structures. The annotations for our microscopy images include seven classes: air, background cells, epidermis, invasive melanoma, inflamed tumor, fibrotic tumor, and uncertain tumor. It is important to note that invasive melanoma is the only type of melanoma precisely annotated. In-situ melanoma is labeled as epidermis in our dataset due to its confinement to the epidermis and due to our focus on segmenting invasive melanoma.

\textbf{Segformer Dataset Generation}. To generate the dataset for Segformer, We follow the approach described by Wang et al \cite{wang2024transformers}. Given the extremely high resolution of microscopy slide images, we divide each slide into non-overlapping patches of $512 \times 512$ and $1024 \times 1024$ pixels. We re-categorize the original annotations into three distinct classes: invasive melanoma, epidermis, and others. This reclassification is essential since the ambiguous boundaries of fibrotic and inflamed tumor make them less suitable for effective segmentation model training. We under-sample the background by discarding patches with 97\% of background cells and air to address class imbalance issues. The resulting dataset comprises 14885 patches of around 3.9 billion pixels for $512 \times 512$ resolution and 4326 patches of around 4.5 billion pixels for $1024 \times 1024$ resolution. This preprocessing approach ensures a balanced and representative dataset for training the Segformer model.

\textbf{SAM Dataset Generation}. We generate the SAM dataset corresponding to the two-stage training process described in \cref{sec:training}. For the first stage, we divide each microscopy image into non-overlapping patches and generate a single point prompt within each connected component randomly. The ground truth for each prompt is the mask corresponding solely to that particular connected component, rather than all components in the patch. This aims to minimize the ambiguity of single point prompts as much as possible and facilitates model convergence. For the second stage, we use both the centroid and randomly sampled points as the training prompts. We use patches centered at each prompt to optimize SAM specifically for our method. This stage's dataset comprises 8357 patches of $512 \times 512$ pixel resolution and 6170 patches of $1024 \times 1024$ pixel resolution.

\subsection{Implementation Details \label{sec:impl}}

\textbf{Model Settings}. We use Segformer B0 and B1 \cite{xie2021segformer} as the initial segmentation model and EfficientSAM-S \cite{xiong2023efficientsam} as the segment anything model in our approach. We choose EfficientSAM-S since it is the most efficient variant of SAM that reconstructs the image embeddings of ViT-H \cite{dosovitskiy2020image} in the original SAM. To further improve training efficiency, we integrate adapters into the ViT \cite{dosovitskiy2020image} image encoder following the approach described in Med-SA \cite{wu2023medical}. We use adapters with an input and output dimensionality of $d_a = 768$ and set the dimensionality of hidden layers to $d_h = 1024$. With adapters, we only fine-tune $6.7\text{M}$ parameters for EfficientSAM-S with a total of $29.3\text{M}$ parameters.

\noindent \textbf{Training}. For Segformer, we use AdamW \cite{loshchilov2017decoupled} optimizer with $\beta_1 = 0.9$ and $\beta_2 = 0.999$, and a weight decay of $0.01$. We use an initial learning rate of $5e-4$ and a polynomial decay scheduler. We use the weights of Segformer pretrained on ImageNet \cite{deng2009imagenet} as our starting point and train the model for 150 epochs. For EfficientSAM, we use the Adam optimizer with an initial learning rate of $1e-4$, an exponential decay rate of $\beta_1 = 0.9$ and $\beta_2 = 0.999$, and a weight decay of $0.05$. We use the initial learning rate for the first $10$ epochs and apply a learning rate decay factor of $0.5$ every 10 epochs. We use a batch size of 8 for patch resolution $512 \times 512$, and a batch size of 6 for patch resolution $1024 \times 1024$. The model is first trained for 100 epochs on the first-stage dataset and then trained for 150 epochs on the second-stage dataset. The model is trained with a binary cross entropy loss function with equal weights for the invasive melanoma and the backgrounds. All experiments are implemented in PyTorch and executed on 4 NVIDIA Quadro RTX 8000 GPUs.

\noindent \textbf{Inference}. To generate the segmentation mask with Segformer, we process each microscopy slide image into patches. Specifically, we use a sliding window to create patches by shifting the window both horizontally and vertically with a step size of $128$ pixels. In addition, we apply a 2D Gaussian kernel as a weighting mechanism for each pixel within a patch. The kernel has the same size as the patch and a standard deviation of $\frac{1}{4}$ of the patch's side length. To generate the segmentation mask for EfficientSAM, we run our method with a fixed set of hyperparameters. We set the threshold for determining in-situ melanoma $\alpha_m = 0.1$, the probability threshold for high-confidence regions $\beta = 0.8$, and the threshold for excluding low-confidence regions $\alpha_c = 0.4$. Additionally, the threshold for the ratio between the sides of a minimum bounding box $\alpha_b$, as described in \cref{alg:prompts}, is set to $3$. The grid prompt employs a vertical and horizontal gap of $64$ pixels between neighboring points.

\subsection{Results \label{sec:exp}}

\begin{table}[t]
\centering
\caption{The results of invasive melanoma segmentation on our dataset. We compare our proposed methods with other state-of-the-art segmentation methods.}
\begin{tabular}{c|c| c c}
\hline
\textbf{Model} & \textbf{Resolution} & \textbf{IoU (\%)} & \textbf{F1 (\%)} \\
\hline
Multi-Scale FCN \cite{phillips2019segmentation} & 512 & 13.0 & 14.0 \\
HRNet \& OCR \cite{shah2023deep} & 788 & 29.1 & 44.0 \\
HIPT \cite{wang2024transformers} & 512 & 40.1 & 57.3 \\
Seg. B0 \cite{xie2021segformer} & 512 & 44.6 & 61.6 \\
Seg. B1 \cite{xie2021segformer} & 512 & 42.0 & 59.2 \\
\hline
Seg. B0 \& EfficientSAM-S (ours) & 512 & \textbf{54.1} & \textbf{70.2} \\
Seg. B1 \& EfficientSAM-S (ours) & 512 & 47.5 & 64.4 \\
\hline \hline
HIPT \cite{wang2024transformers} & 1024 & 33.0 & 46.0 \\
Seg. B0 \cite{xie2021segformer} & 1024 & 44.0 & 61.1 \\
Seg. B1 \cite{xie2021segformer} & 1024 & 45.0 & 62.0 \\
\hline
Seg. B0 \& EfficientSAM-S (ours) & 1024 & \textbf{49.5} & \textbf{66.2} \\
Seg. B1 \& EfficientSAM-S (ours) & 1024 & 49.4 & 66.1 \\
\hline
\end{tabular}
\label{tab:cp}
\end{table}

\begin{figure}[t]
\captionsetup[subfigure]{justification=Centering}

\begin{subfigure}[t]{0.3\textwidth}
    \centering
    \includegraphics[width=\linewidth]{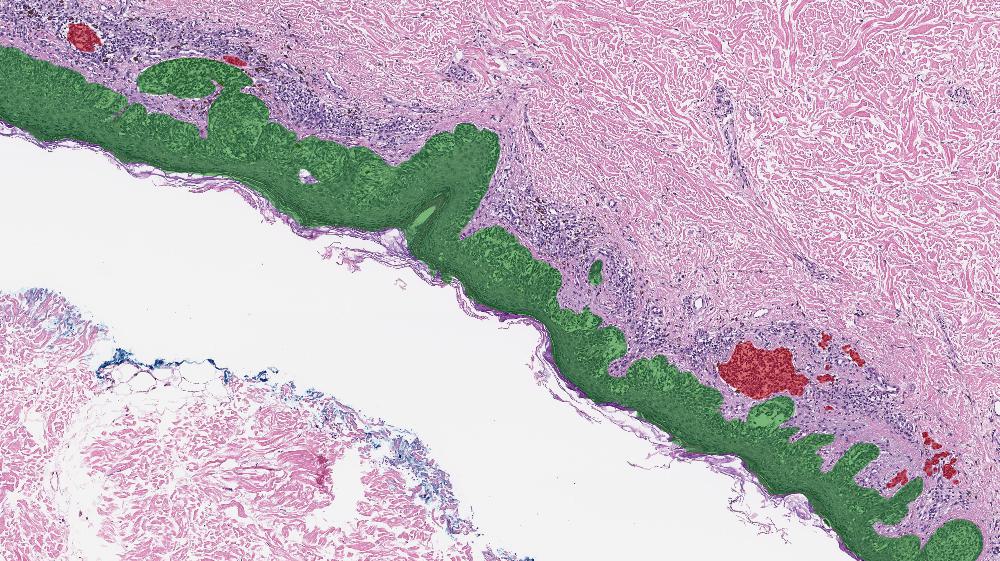} 
\end{subfigure}
\centering
\begin{subfigure}[t]{0.3\textwidth}
    \centering
    \includegraphics[width=\linewidth]{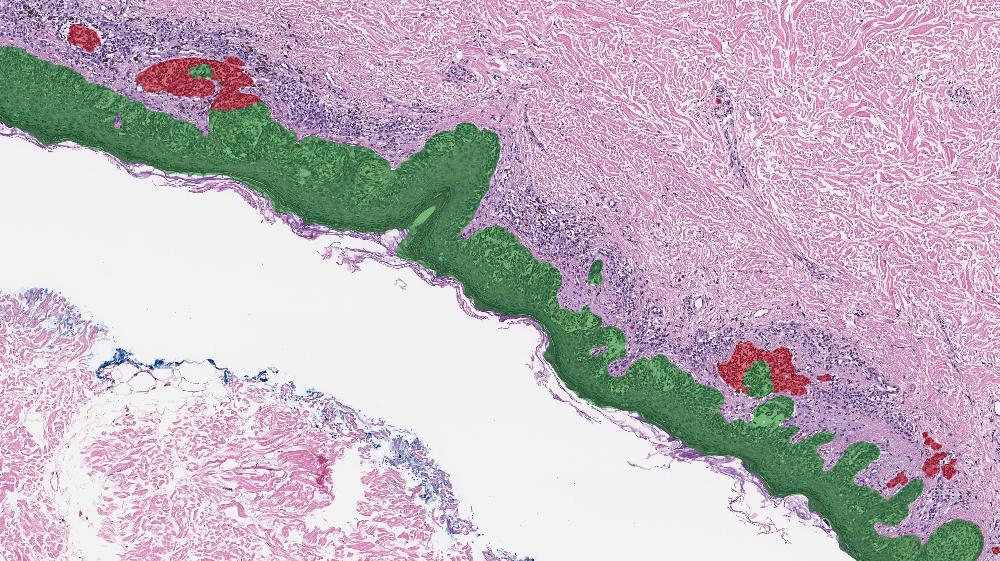}
\end{subfigure}
\begin{subfigure}[t]{0.3\textwidth}
    \centering
    \includegraphics[width=\linewidth]{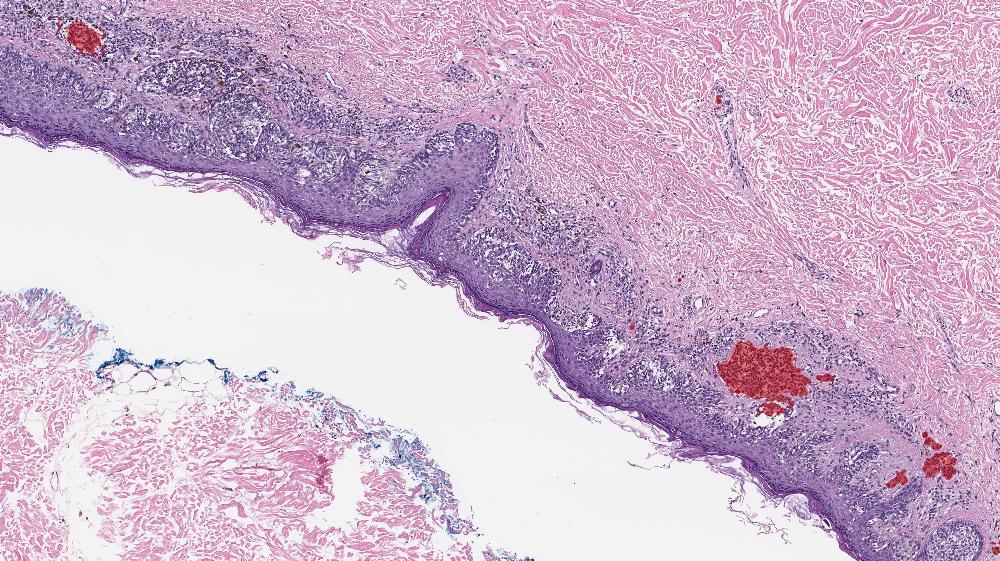}
\end{subfigure}

\medskip
\begin{subfigure}[t]{0.3\textwidth}
    \centering
    \includegraphics[width=\linewidth]{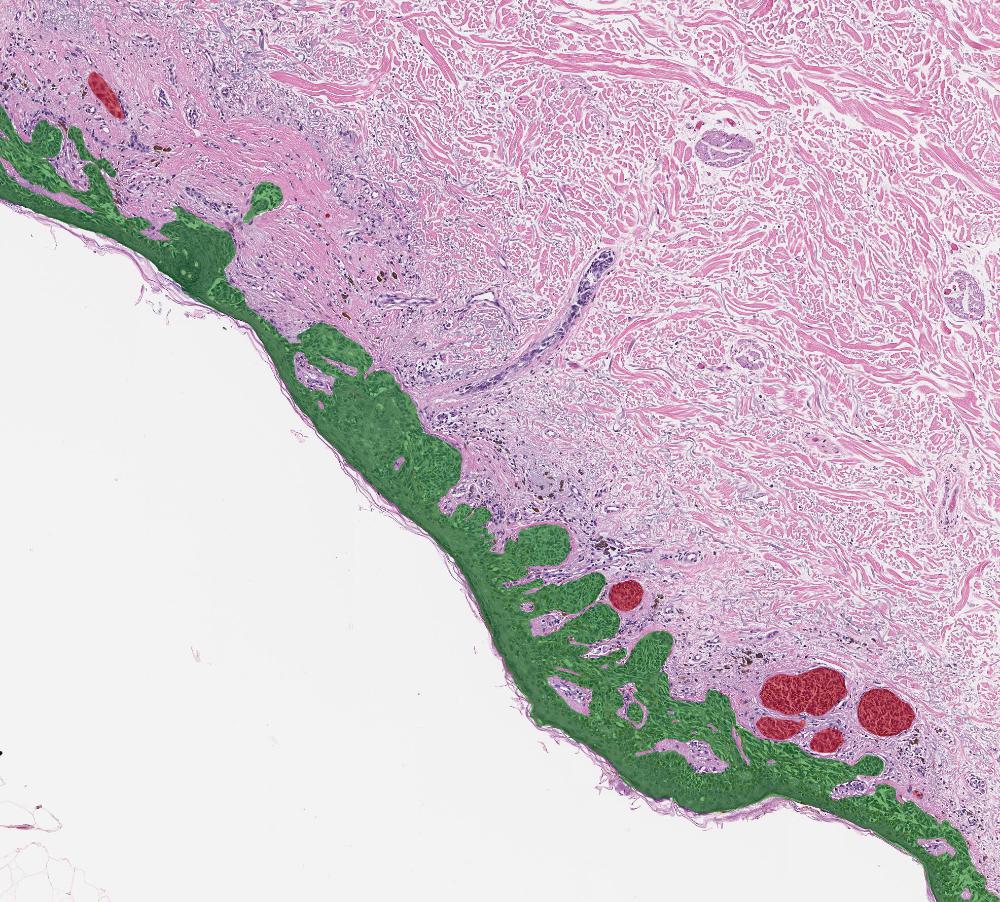}
\end{subfigure}
\centering
\begin{subfigure}[t]{0.3\textwidth}
    \centering
    \includegraphics[width=\linewidth]{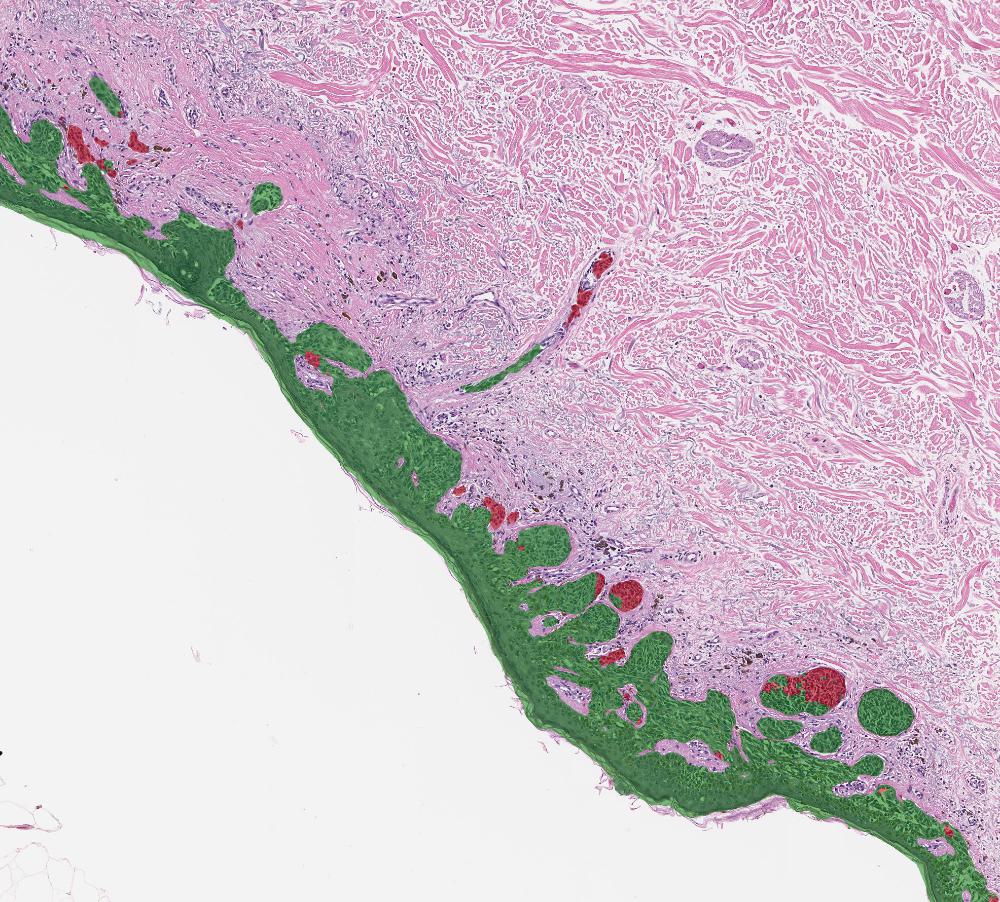}
\end{subfigure}
\begin{subfigure}[t]{0.3\textwidth}
    \centering
    \includegraphics[width=\linewidth]{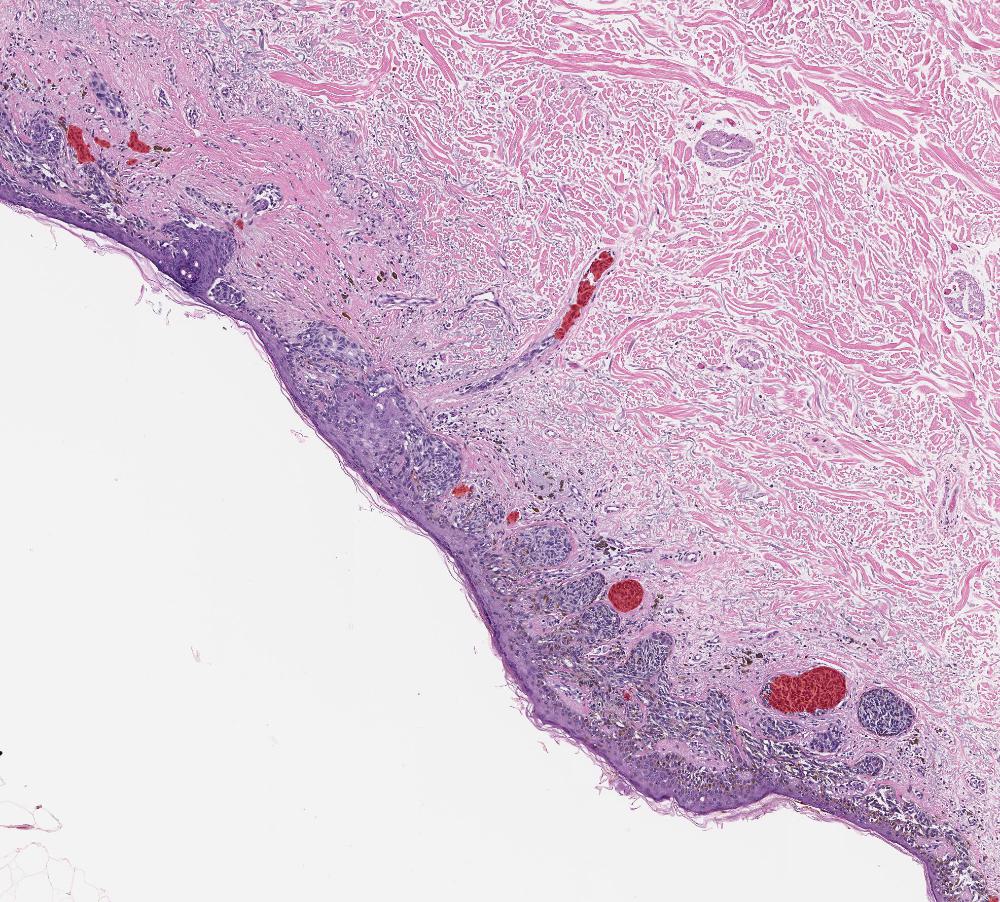}
\end{subfigure}

\medskip
\begin{subfigure}[t]{0.3\textwidth}
    \centering
    \includegraphics[width=\linewidth]{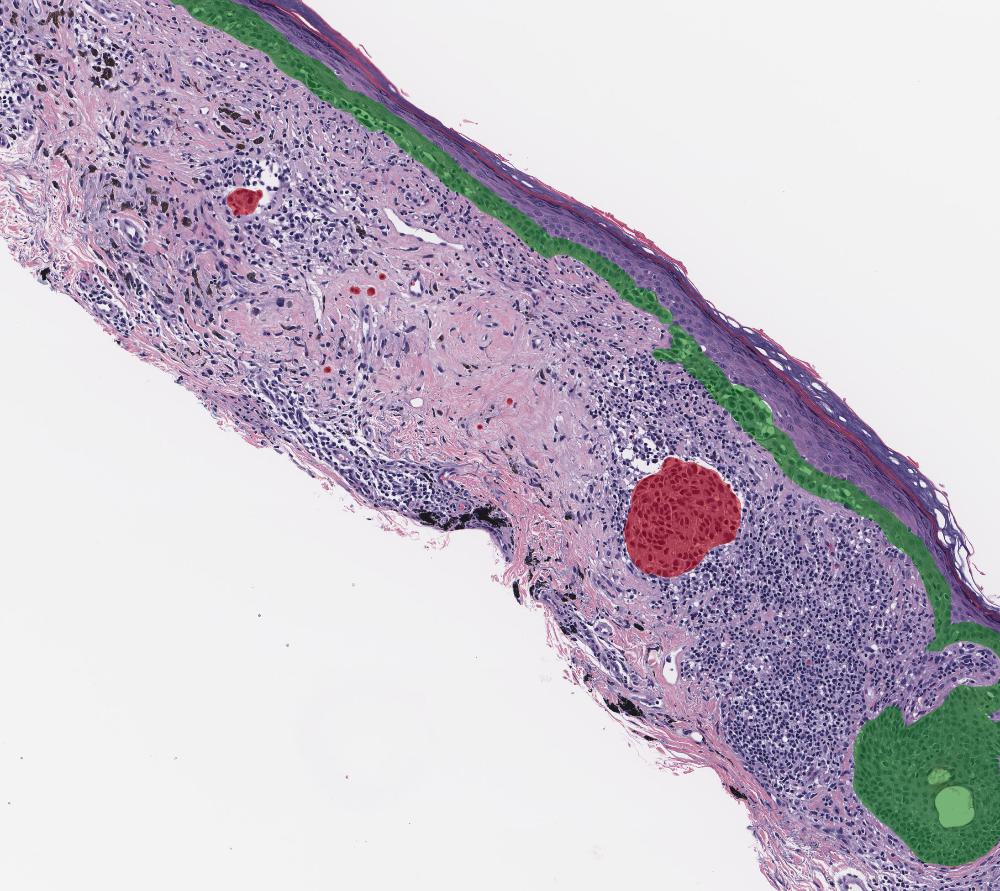}
    \caption{Ground Truth}
\end{subfigure}
\centering
\begin{subfigure}[t]{0.3\textwidth}
    \centering
    \includegraphics[width=\linewidth]{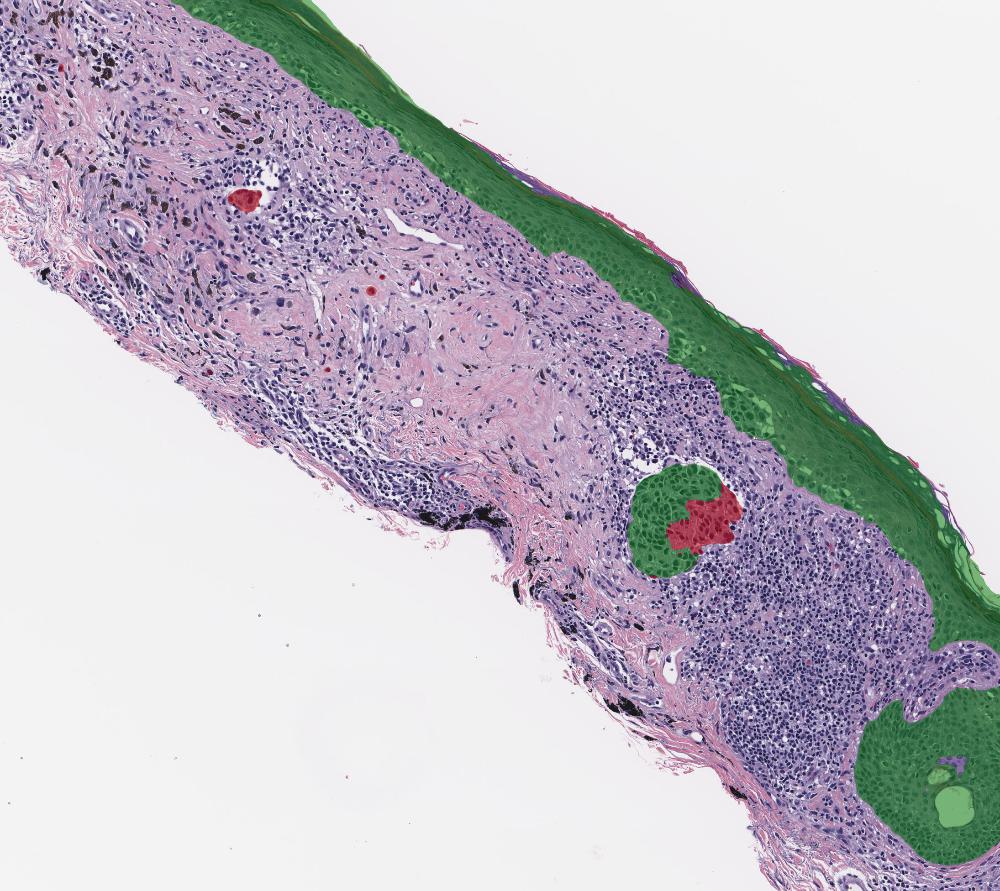}
    \caption{Segformer-B0}
\end{subfigure}
\begin{subfigure}[t]{0.3\textwidth}
    \centering
    \includegraphics[width=\linewidth]{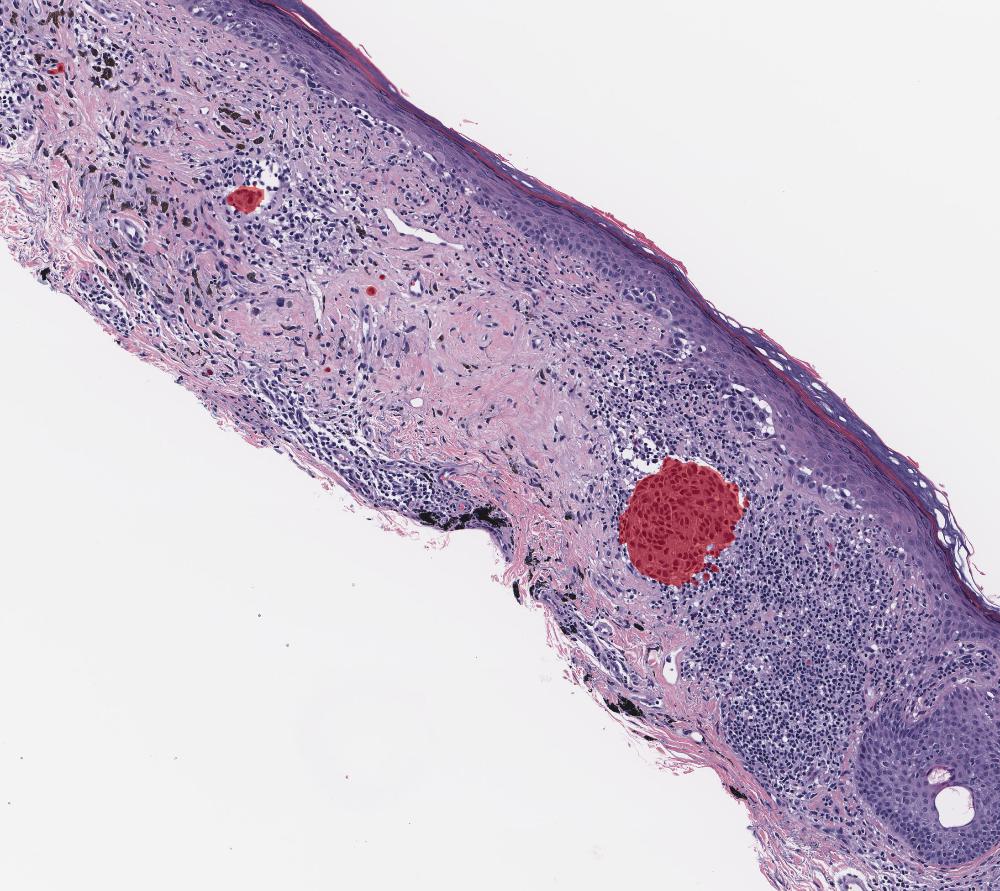}
    \caption{Ours}
\end{subfigure}
\caption{Qualitative results on our dataset. Red denotes invasive melanoma and green denotes the epidermis. Compared to Segformer, our method significantly improves the accuracy of predictions for invasive melanoma regions, especially in areas where distinguishing from the epidermis is challenging.}
\label{fig:vis}
\end{figure}

To evaluate the effectiveness of our proposed method, we compare it with state-of-the-art melanoma segmentation methods on our dataset. As shown in \cref{tab:cp}, we compare our method with melanoma segmentation methods including Multi-Scale FCN \cite{phillips2019segmentation}, HRNet \& OCR \cite{shah2023deep}, HIPT \cite{wang2024transformers}, and Segformer \cite{xie2021segformer}. The results for Segformer B0 and B1 are reproduced following the methods described in Wang et al \cite{wang2024transformers}. All EfficientSAM methods shown in the table use a $1024 \times 1024$ patch size, which is different from the resolution field in \cref{tab:cp} that represents the side length of a patch used for the initial segmentation model. 

As shown in \cref{tab:cp}, our method outperforms all other baselines in terms of IoU and F1 score. Compared to using solely the Segformer, which is also used as the initial segmentation model in our method, our method achieves a gain of 9.1\% IoU and 8.2\% F1 over the state-of-the-art segmentation method. \cref{fig:vis} presents representative qualitative results on our dataset, where our method significantly reduces errors in in-situ melanoma regions and enhances segmentation accuracy in areas where distinguishing between invasive melanoma and the epidermis is particularly challenging. Additionally, the best result is achieved with using EfficientSAM and Segformer B0 with $512 \times 512$ patches, even though this is not the best performing Segformer in terms of IoU and F1 score. A higher IoU for the initial segmentation mask does not necessarily indicate a higher improvement in IoU in the final mask of our method. The amount of improvement depends on the quality of sampled prompts and the room left for improvement for each prompt.

\subsection{Ablation Studies \label{sec:ablation}}

We now analyze our method through a series of ablation studies on each component of the framework.

\subsubsection{In-situ Melanoma Detection}

\begin{table}[tb]
\centering
\caption{Ablation study on in-situ melanoma detection and low-confidence region filtering in preparation for prompt generation, as described in \cref{sec:detection} and \cref{sec:filter}.}
\begin{tabular}{c|c|c c|c}
\hline
\begin{tabular}{@{}c@{}}\textbf{Initial} \\ \textbf{Model} \end{tabular} & \textbf{Resolution} & \begin{tabular}{@{}c@{}}\textbf{In-situ Melanoma} \\ \textbf{Detection}\end{tabular} & \begin{tabular}{@{}c@{}}\textbf{Low-Confidence} \\ \textbf{Region Filtering}\end{tabular} & \textbf{IoU (\%)} \\
\hline
Seg. B0 & 512 & & & 42.4 \\
Seg. B0 & 512 & \checkmark & & \textbf{54.2} \\
Seg. B0 & 512 & \checkmark & \checkmark & 54.1 \\
\hline
Seg. B0 & 1024 & & & 41.4 \\
Seg. B0 & 1024 & \checkmark & & 45.6 \\
Seg. B0 & 1024 & \checkmark & \checkmark & \textbf{49.5} \\
\hline
Seg. B1 & 512 & & & 44.1 \\
Seg. B1 & 512 & \checkmark & & 47.5 \\
Seg. B1 & 512 & \checkmark & \checkmark & \textbf{47.5} \\
\hline
Seg. B1 & 1024 & & & 44.3 \\
Seg. B1 & 1024 & \checkmark & & 44.8 \\
Seg. B1 & 1024 & \checkmark & \checkmark & \textbf{49.4} \\
\hline
\end{tabular}
\label{tab:in-situ}
\end{table}

We investigate the impact of in-situ melanoma detection described in \cref{sec:detection} by filtering out all detected estimated in-situ melanoma regions regardless of their confidence levels. \cref{tab:in-situ} shows that this consistently improves performance, with IoU gains ranging from 0.5\% to 11.8\%. The improvement varies based on the quality of the initial segmentation mask and the room left for improvement. In contrast, disabling in-situ melanoma detection results in a performance drop compared to the original mask produced by Segformer. This drop occurs because prompts generated from incorrect predictions lead SAM to make additional errors. Therefore, enabling in-situ melanoma detection not only mitigates errors from incorrect initial predictions but also leverages the strengths of the remaining high-confidence predictions to improve the final segmentation results.

\subsubsection{Low-Confidence Region Filtering}

We study the impact of low-confidence region filtering presented in \cref{sec:filter}. As shown in \cref{tab:in-situ}, our method achieves a 3.9\% IoU gain for Segformer B0 and a 4.6\% IoU gain for Segformer B1 when using a $1024 \times 1024$ patch size. However, there is a slight drop of 0.1\% in IoU for Segformer B0 with a $512 \times 512$ patch size. This suggests that some high-confidence invasive melanoma regions touch the epidermis, which are not filtered by the algorithm, are actually in-situ melanoma. Overall, this demonstrates that low-confidence region filtering improves the robustness of our segmentation results.

\subsubsection{Prompt Types}

\begin{table}[tb]
\centering
\caption{Ablation study on the effects of using different prompts. "Both" denotes the method that dynamically uses both centroid and grid prompts based on the shape distributions of melanoma components, as shown in \cref{sec:prompt}.}
\begin{tabular}{@{} c | c | c | c c @{}}
\hline
\begin{tabular}{@{}c@{}}\textbf{Initial} \\ \textbf{Segmentaion Model} \end{tabular} & \textbf{Prompt} & \textbf{Resolution} & \begin{tabular}{@{}c@{}}\textbf{EfficientSAM} \\ \textbf{IoU} (\%) \end{tabular} & \begin{tabular}{@{}c@{}}\textbf{Final Mask} \\ \textbf{IoU (\%)} \end{tabular} \\
\hline
Seg. B0 & Centroid & 512 & 43.3 & 53.1 \\
Seg. B0 & Grid & 512 & 50.6 & 51.7 \\
Seg. B0 & Both & 512 & \textbf{52.2} & \textbf{54.1} \\
\hline
Seg. B1 & Centroid & 512 & 44.3 & 47.5 \\
Seg. B1 & Grid & 512 & \textbf{46.4} & 46.5 \\
Seg. B1 & Both & 512 & 46.1 & \textbf{47.5} \\
\hline
Seg. B0 & Centroid & 1024 & 40.8 & 48.5 \\
Seg. B0 & Grid & 1024 & 45.5 & 45.8 \\
Seg. B0 & Both & 1024 & \textbf{48.7} & \textbf{49.5} \\
\hline
Seg. B1 & Centroid & 1024 & 37.7 & 47.3 \\
Seg. B1 & Grid & 1024 & 46.0 & 47.2 \\
Seg. B1 & Both & 1024 & \textbf{47.1} & \textbf{49.4} \\
\hline
\end{tabular}
\label{tab:prompt}
\end{table}

We study the effectiveness of different prompt types. We test prompts with centroid alone, grid alone and our proposed strategy that uses both as presented in \cref{sec:prompt}. \cref{tab:prompt} shows that in most cases using both achieves the highest IoU in the final mask. Compared to grid prompts alone, using both achieves a gain as high as 3.7\% IoU. Compared to centroid prompts alone, using both achieves gains up to 2.1\% IoU with one case showing no improvement. It is noticeable that centroid prompts alone outperforms grid prompts alone in the final mask, but performs much worse in pre-merged mask. This shows that centroid prompts allow more accurate segmentation and thus complement the initial mask well, achieving high IoU in the final mask after merging. In contrast, grid prompts alone achieve full coverage over the initial mask, but not all points serve as effective prompts for accurate segmentation.  This demonstrates that our method effectively leverages the strengths of both prompts by using grid prompts for large, irregular melanoma components and centroid prompts for small, regular melanoma components, maximizing the advantages of each prompt type.

\begin{table}[tb]
\centering
\caption{Ablation study on final mask generation. The last row represents the result when prompts are generated from the ground truth instead of the mask produced by Segformer.}
\begin{tabular}{@{} c | c | c c c @{}}
\hline
\begin{tabular}{@{}c@{}}\textbf{Initial} \\ \textbf{Segmentaion Model} \end{tabular} & \textbf{Resolution} & \begin{tabular}{@{}c@{}}\textbf{Post-processed} \\ \textbf{IoU (\%)}\end{tabular} & \begin{tabular}{@{}c@{}}\textbf{EfficientSAM} \\ \textbf{IoU (\%)}\end{tabular} & \begin{tabular}{@{}c@{}}\textbf{Final} \\ \textbf{IoU} (\%)\end{tabular} \\
\hline
Seg. B0 & 512 & 48.6 & 52.2 & \textbf{54.1} \\
Seg. B1 & 512 & 43.4 & 46.1 & \textbf{47.5} \\
Seg. B0 & 1024 & 46.7 & 48.7 & \textbf{49.5} \\
Seg. B1 & 1024 & 47.2 & 47.1 & \textbf{49.4} \\
\hline
GT & - & - & 63.0 & - \\
\hline
\end{tabular}
\label{tab:final}
\end{table}

\subsubsection{Final Mask Generation}

We study the impact of final mask generation. As shown in \cref{tab:final}, the final mask consistently achieves the highest IoU in all cases. EfficientSAM's mask outperforms the post-prococessed mask in most cases, with the most significant improvement being a 5.5\% increase for Segformer B0 using $512 \times 512$ patches. Merging two masks into a final mask significantly enhances accuracy. The post-processed mask shows gains ranging from 2.2\% to 5.5\% IoU, while the EfficientSAM's mask gains improvements between 0.8\% and 2.3\% IoU. This demonstrates that EfficientSAM compliments well the post-processsed mask in generating the final mask. In addition, we evaluate the EfficientSAM's performance with prompts generated from the ground truth. Its IoU performance is 10.8\% higher than the best result of using EfficientSAM alone and 8.9\% higher than the best final mask. This demonstrates the upper limit of the performance of our method when using fully accurate prompts.

\section{Conclusion}
We proposed a novel approach to explore the potential of SAM for melanoma segmentation in microscopy slide images. Our method utilizes Segformer to generate initial segmentation masks and subsequently prompts EfficientSAM using a dynamic selection of centroid and grid prompts for automatic invasive melanoma segmentation. Our experimental results demonstrate that this approach surpasses other state-of-the-art melanoma segmentation methods by a large margin of 9.1\% in IoU.

\section*{Acknowledgements}

We would like to express our gratitude to Michael Wang and Timothy McCalmont for providing the data, and to Shinwoo Choi for sharing valuable insights, without which this paper would not be possible.

\bibliographystyle{splncs04}
\bibliography{main}
\end{document}